\documentclass{article}
\usepackage{spconf,amsmath,graphicx}
\usepackage{xcolor}
\usepackage{multirow}
\newcommand\linesubsec[1]{\vspace{0.8mm}\noindent\textbf{#1}}
\usepackage{subcaption,siunitx,booktabs}
\usepackage[margin=1in]{geometry}
\usepackage{multirow}

\title{Rank-based loss for learning hierarchical representations}
\name{In\^{e}s Nolasco$^{1}$, Dan Stowell$^{2}$ }%\thanks{Thanks to XYZ agency for funding.}}
\address{$^1$ Centre for Digital Music (C4DM), Queen Mary University of London, London, UK\\$^2$ Tilburg University, Tilburg, The Netherlands; Naturalis Biodiversity Centre, Leiden, The Netherlands\\}
\begin{document}
%\ninept
%
\maketitle
\begin{abstract}
Hierarchical taxonomies are common in many contexts, and they are a very natural structure humans use to organise information. 
In machine learning, the family of methods that use this "extra" information is called hierarchical classification. However, applied to audio classification, this remains relatively unexplored.
Here we focus on how to integrate the hierarchical information of a problem to learn embeddings representative of the hierarchical relationships. Previously, triplet loss has been proposed to address this problem, however it presents some issues like requiring the careful construction of the triplets, and being limited in the extent of hierarchical information it uses at each iteration.
In this work we propose a rank based loss function that uses hierarchical information and translates this into a rank ordering of target distances between the examples. We show that rank based loss is suitable to learn hierarchical representations of the data. By testing on unseen fine level classes we show that this method is also capable of learning hierarchically correct representations of the new classes. 
Rank based loss has two promising aspects, it is generalisable to hierarchies with any number of levels, and is capable of dealing with data with incomplete hierarchical labels.

% \textcolor{red}{ TODO- Finish follow nature example! :for that we propose a novel loss function -rank based loss ...
% we explore its feasibility in different settings and datasets and also compare it with a state of the art approach.\\
% Dan says: mention the triplet loss and either their potential inefficiency or that they are an indirect way to train a hierarchical representation.}

\end{abstract}
\begin{keywords}
hierarchical data, metric learning, rank, embeddings
\end{keywords}
%

% \vspace{-0.3cm}
\section{Introduction}
\label{sec:intro}

Many supervised learning tasks can be framed as hierarchical problems, meaning that the taxonomy that organises the label space can be constructed  as to follow a hierarchical tree structure. Instead of having a flat single level label space, in a hierarchical tree structure the labels are organised in different levels and there is a hierarchical relationship between them.
Two important characteristics of these taxonomies are that each child label only has one parent, and it is expected that children from the same parent to share a closer similarity (concept and features) with each other than with labels from other parents. Furthermore this similarity increases as we go down in the hierarchy towards the leaf-labels.
%To give an example from bioacoustics, in Fig.\ref{fig:aaii_taxonomy}  we define a label structure based on animal taxonomy. Going from the taxonomic group at the higher level, to species at the intermediate level and finally the individual animal identification at the leaf level. 
% \begin{figure}[t]
% \centering
% \includegraphics[width=0.5\textwidth]{hierarchical_label_tree.png}
% \caption{Example of a hierarchical taxonomy for the task of individual animal identification. This taxonomy shows 3 levels of labels, The higher level is the taxonomic group - bird, mammal; the intermediate level is the species, and  leaf-level is the id of the individual animal.}
% \label{fig:aaii_taxonomy}
% \end{figure}

Hierarchical classification is supervised learning that makes use of this kind of label structure to not only make better predictions at the leaf-level task, but also to generate predictions at all the other levels of the taxonomy. This approach is attractive in the sense that we build models that give us a complete picture of how the objects are organised and how they are related with other classes. Furthermore it offers the possibility of classifying items of unknown "leaf" classes into a broad category.

In this work we propose a novel loss function that uses the partial ordering implied by the hierarchical label taxonomy to derive target distances between embeddings and thus learn hierarchical meaningful embeddings, \textit{i.e.,} embeddings that represent the hierarchical relationships in the taxonomy. 
Unlike other methods, rank based Loss (RbL) is suitable to learn hierarchical taxonomies with any number of levels and conceptually it is also appropriate to deal with examples with incomplete labels, where we may know higher order classes but may be missing the more specific classes. 
% In this work, we mainly explore how this loss function compares with the quadruplet loss approach proposed in \cite{jati_hierarchy-aware_2019}, and how it behaves with different datasets.

% CONTRIBUTIONS
Our main contributions are: 1) the rank-based loss and 2) preliminary analysis on how RbL behaves in different settings and datasets and how this compares to a quadruplet loss function that integrates hierarchical information proposed in \cite{jati_hierarchy-aware_2019}.
% paper organization: 
The remaining of this paper is organised as: Related work in sec.\ref{sec:related_work}, in sec.\ref{sec:methods} the proposed rank based loss is defined, followed by the description of the datasets, feature extraction and network architecture used,  evaluation approach and finally experiments. In Sec.\ref{sec:res} we present the results and a discussion. Section \ref{sec:conc} we conclude this work with a reflection for future directions and improvements.

% \vspace{-0.2cm}

\section{RELATED WORK}
\label{sec:related_work}

While the concept of hierarchical classification is not new, the application to audio data and within a deep learning framework has not been fully explored.
The works in \cite{pham_deep_2020} and \cite{ganchev2007automatic} employ the classical method of hierarchical classification based on training separate models for each level of the hierarchy. The predictions at the fine-level classes are simply the result of combining the prediction probabilities at the higher levels. 

From the literature, current non-classical approaches to hierarchical classification are mainly two:
Multi-task learning, in which we have one task per level of the hierarchy and following the multi-task learning framework, the network learns to optimise for the multiple tasks together.  In this vein, the work in \cite{xu_hierarchical_2016} deals with acoustic scene classification on a 2 level tree label structure. Besides designing an objective function that combines together the loss at the fine level and coarse level, the authors also propose a scheme of pretraining the networks on a single level of the label tree in order to improve the training and performance on other levels.
Another pertinent example is the work in \cite{cramer_chirping_2020} that focuses on bird species classification from flight calls. Here the data is organised in a 3 level taxonomy (animal order, family, species). The authors propose a novel network architecture, the \textit{Taxonet} that is both a hierarchical and multi-task neural network. By partitioning the layers and defining conditional activation of each node given which partition was activated in the previous layer, they are able to translate the hierarchical taxonomy of the problem into the network architecture.
Both these examples report improvements using their hierarchical multi-task methods when compared against flat classification, \textit{i.e,.} classification only at the fine level.
The other main approach for hierarchical classification is more related with metric learning and generating embeddings that explicitly convey the hierarchical structure of the problem. The core idea is that distances between similarly labelled examples should be minimised and examples more distant in the label space should have their distances maximised. 
With this intent, in \cite{elizalde2018sound}, the authors explore the use of Siamese networks and manually define a target distance between pairs of items in the generated embeddings depending on the position of the input examples on the label tree. Hierarchical information is also integrated through the network architecture by multiplying an incidence matrix with the output layer predicting the leaf-level of classes which generated the output predictions for the higher level classes. 
In \cite{garcia_leveraging_2021}, the authors address musical instrument recognition in a few shot setting. They use prototypical networks to generate embeddings at both levels of a music instrument hierarchy, by aggregating the leaf-level embeddings according to the label structure of the problem.
Another relevant work \cite{jati_hierarchy-aware_2019} employs a quadruplet loss (generalisation of triplet loss) approach to sound event detection on a dataset organised in a two level hierarchical tree. The core of their proposed method is to build quadruplets that contain examples of all the possible hierarchical relationships of the label tree. \textit{I.e.,} anchor and positive are examples from the same leaf-label, negative are examples from the same coarse level and different leaf-label, or from a different coarse level.
The authors report improved classification results at both levels.
% hierarchical classification, and learning embeddings by  using the hierarchical structure of the labels.
% distance metric learning.
%     hierarchical classification papers from presentation

% rank learning related work
Loosely related with learning hierarchical relationships is also the concept of learning to rank. In learning to rank the goal is to learn a function that given a query example will score relevant/closer results higher than irrelevant ones or simply sort a list of results by the relevance to the query.
In \cite{mcfee_metric_nodate} the authors frame the problem of learning to rank as a metric learning problem, and propose a method that directly optimises for ranking.
% Following this, the authors in \cite{deep_metric_learning_rank} explore ...
% rank learning applied to hierarchical classification?
Also, in \cite{naik_ranking-based_2015} the concept of rank is explored connected to a classical hierarchical classification approach: the authors propose the training of leaf-level classifiers weighting more data examples that are closest in ranking to each class being trained.

% \vspace{-0.4cm}

\section{Methods}
\label{sec:methods}

\subsection{Rank-based loss}
\label{subsec:rbl}

% \begin{figure}[t]
% \centering
% \includegraphics[width=0.5\textwidth]{compute_RBL_example.png}
% \caption{\textcolor{red}{CORRECT FIG AND MAKE IT LARGER! or just remove it for space} Example for computing Loss for one batch.}
% \label{fig:compute_rbl_example}
% \end{figure}
% % \subsection{Quadruplet Loss}
% % The Quadruplet loss was proposed in \cite{jati_hierarchy-aware_2019}

Our proposed loss function\footnote{The Pytorch implementation of Rank based loss is available:  https://github.com/inesnolas/Rank-based-loss\_ICASSP22} follows a metric learning approach, where the objective is to learn embeddings in which the distances between them are meaningful to the problem being addressed. Similarly to the quadruplet loss proposed in \cite{jati_hierarchy-aware_2019}, at each iteration of training we want to evaluate the distances between embeddings and push the embeddings closer or further away depending on the hierarchical relationship between labels. 
The hierarchical information is used here to define, for each element, the desired rank-ordering of all other elements in terms of their distance.
For each pair of embeddings we compute a loss value that is either 0 if the pair has a "correct" distance given the rank, or a positive value meaning how far away from the target distance the pair is.
Formally the rank based loss is defined as:
\begin{equation}
    L =  \frac{1}{P}\sum_{p}^{P}(1-I_{p}).(EmbDist_{p} - TargetDist_{p})^{2}
    \label{eq:RbL}
\end{equation}
where $EmbDist_{p}$ is the cosine distance in the embedding space between two embeddings. $TargetDist_{p}$ is the target distance for that pair given the desired rank of the pair. $I_{p}$ is a Boolean indicating if the pair is correctly distanced given their rank in the label tree.

We summarise computation of this loss in 5 steps:%, and a toy example of computing the loss for one batch of data is shown in Fig. \ref{fig:compute_rbl_example}
% \begin{enumerate}
%     \item Compute a rank map from the tree of ground truth labels: Each pair of examples has a rank given by the tree distance of their labels. The tree distance is given by the number of nodes that separate the two labels.
%     \item Compute all the pairwise cosine distances in the batch in the embedding space, and sort them.
%     \item For each rank, assign a target distance by selecting whatever distance in the sorted distances vector falls at each rank.
%     \item Compute $I_{p}$ as: $0$ if distance of the pair is within the correct positions in the sorted distances vector, else $1$ if distance of the pair is wrong given the ground truth rank.
%     \item Compute the loss from eq.\ref{eq:RbL}.
% \end{enumerate}

\linesubsec{1)} Compute a rank map from the tree of ground truth labels: Each pair of examples has a rank given by the tree distance of their labels. The tree distance is given by the number of nodes that separate the two labels

\linesubsec{2)} Compute all the pairwise cosine distances in the batch in the embedding space, and sort them.

\linesubsec{3)} For each rank, assign a target distance by selecting whatever distance in the sorted distances vector falls at each rank.

\linesubsec{4)} Compute $I_{p}$ as: $0$ if distance of the pair is within the correct positions in the sorted distances vector, else $1$ if distance of the pair is wrong given the ground truth rank.

\linesubsec{5)} Compute the loss from eq.\ref{eq:RbL}.

% \textcolor{red}{TODO: justify here our design for the loss function}
% \vspace{ -0.5cm}
\subsection{Datasets}
\label{subsec:data}
Three datasets from different contexts were used:
% \begin{description}
% \item
% \end{description}

\linesubsec{3 Bird species,} audio data collected to accompany the work in \cite{stowell_automatic_2019} on automatic acoustic identification of individual animals. It contains labelled recordings of individuals from three different bird species: Little owl (Athene noctua), Chiffchaff (Phylloscopus collybita), and Tree pipit (Anthus trivialis). For this work, the data was augmented by mixing foreground recordings of each individual with background recordings of other individuals. Furthermore we re-structured the labels to follow a 3 level hierarchical taxonomy:  taxonomic group, species, and individual identity. From this dataset, a total of 1707 recordings were selected belonging to 9 individuals equally distributed across the 3 species of birds.
    
\linesubsec{Nsynth,} a large-scale dataset of annotated musical notes \cite{nsynth2017}, it contains 4-second audio snippets of notes played with different instruments. For this work a small selection of the dataset was created to address the task of instrument recognition. A 2-level hierarchical taxonomy is built using the instrument family labels as highest level classes, and the instrument id as the fine-level. We selected a total of 1707 audio snippets from 9 instruments across the guitar, flutes and keyboard families. All instruments are from the "acoustic" source. %A second dataset was created containing a total of 39 instruments across the same 3 families.
    
\linesubsec{TUTasc2016,} dataset of 30-second audio segments from 15 acoustic scenes \cite{tut_asc_2016_data}. These are organised in 3 groups: indoor, outdoor and vehicles accordingly to the environment where they were captured. For the hierarchical taxonomy we consider the acoustic scene labels as the fine level classes and the 3 groups as the coarse level classes. The selected data used in this work consists of 704 recordings coming from 9 acoustic scenes, balanced across the 3 groups: library, home, and metro station from indoors; tram, bus and train from vehicles; residential area, forest path and beach from outdoors.

\begin{table*}[htp]
%  \resizebox{\textwidth}{!}{%
\centering
% \begin{subtable}[c]{\textwidth}
% \centering
\resizebox{\textwidth}{!}{  
\begin{tabular}{c|l|c|c|c||c|c|c||c|c|c}

 & & \multicolumn{3}{c||}{\textbf{3 bird species}} & \multicolumn{3}{c||}{\textbf{Nsynth}} & \multicolumn{3}{c}{\textbf{TUTasc2016}} \\ 
 & & $Sil^{Fine}$   & $Sil^{Coarse}$ & $avSil$  & $Sil^{Fine}$   &$Sil^{Coarse}$ & $avSil$  & $Sil^{Fine}$   & $Sil^{Coarse}$ &$avSil$ \\ \hline
   \parbox[t]{2mm}{\multirow{4}{*}{\rotatebox[origin=c]{90}{Test}}} 
&\textbf{InitEmb}&\textbf{	-0.07 (0.0)} &0.19 (0.0) &	0.06 (0.0)      &    \textbf{0.02 (0.0)}	&	0.29 (0.0)   &		0.16 (0.0)                         &  -0.02 (0.0)&		0.18 (0.0)	&	0.08 (0.0) \\
&\textbf{QuadL} &	-0.17 (-0.10)&	0.32 (+0.13)&	0.07 (+0.01)&	 0.01 (-0.01)&\textbf{	0.60 (+0.31)}& \textbf{0.31 (+0.15)}&  	-0.19 (-0.17)&	0.14 (-0.04)&	-0.02 (-0.10)\\
&\textbf{RbL} &	-0.09 (-0.02)&	\textbf{0.42 (+0.23)}& \textbf{0.17 (+0.11)}&	   -0.08 (-0.10) &	0.46 (+0.17) &	0.19 (+0.03)&            0.03 (+0.05)&	0.60 (+0.42)&	0.31 (+0.23)\\
&\textbf{RbL\_unc }&	-0.23 (-0.16)&	0.23 (+0.04)&0.0 (-0.06)	&   -0.16 (-0.18)&	0.38 (+0.09)&	0.11 (-0.05)&   \textbf{0.06 (+0.08)}&	\textbf{0.73 (+0.55)}&	\textbf{0.40 (+0.32)}\\
% RbL_RdmHierarchy &	-0.15 (-0.16)&	-0.09 (-0.3)&	-0.12 (-0.23)&	 0.04 (-0.11)&	0.41 (+0.11)&	0.23 (+0.01)& 	-0.23 (-0.44)&	-0.04 (-0.30)&	-0.14 (-0.38) \\

  \hline
  \hline
     \parbox[t]{2mm}{\multirow{4}{*}{\rotatebox[origin=c]{90}{NovelTest}}} 
&\textbf{InitEmb}&\textbf{-0.06 (0.0) }&0.35 (0.0)&0.14 (0.0)&\textbf{-0.02 (0.0)}&0.04 (0.0)&0.01 (0.0)&0.17 (0.0)&0.22 (0.0)&0.19 (0.0)\\
&\textbf{QuadL}&-0.08 (-0.02)&\textbf{0.57 (+0.22)}&	\textbf{0.25 (+0.11)}& 	-0.09 (-0.07)&	-0.02 (-0.06)&	-0.05 (-0.06)& 0.14 (-0.03)&	0.27 (+0.05)&	0.21 (+0.02)\\
&\textbf{RbL}&\textbf{	-0.06 (0.0)}&	0.48 (+0.13)&	0.21 (+0.07)&           	-0.04 (-0.02)&	\textbf{0.13 (+0.09)}&	\textbf{0.04 (+0.03)}&                 0.16 (-0.01) &\textbf{	0.8 (+0.58)} &	0.48 (+0.29)\\
&\textbf{RbL\_unc}&	-0.19 (-0.13)&	0.32 (-0.03)&	0.07 (-0.07)                &  -0.33 (-0.31)&	0.06 (+0.02)&	-0.13 (-0.14)&\textbf{ 0.33 (+0.16)}&	0.74 (+0.52)&\textbf{	0.53 (+0.34)}\\
% RbL_RdmHierarchy&	-0.16 (-0.33)&	-0.12 (-0.51)&	-0.14 (-0.42)&  	-0.1 (-0.28)&	-0.12 (-0.26)&	-0.11 (-0.27)& 	-0.02 (-0.49)&	0.04 (-0.35)&	0.01 (-0.42) \\

\end{tabular}%
}

% \end{subtable}
\caption{ Silhouette scores on the test sets and on the alternative test sets, where the leaf-classes are different from the classes used to train the models.}
\label{tab:res}
\end{table*}

\subsection{Feature extraction and architecture}
\label{subsec:feat}
% The focus of this work is to analyse the use of the proposed loss function RbL in different setups. Given that, we devised a simple feature extraction system and network intended to give a good enough basis for all the experiments.
% # Architectural constants.
% NUM_FRAMES = 96  # Frames in input mel-spectrogram patch.
% NUM_BANDS = 64  # Frequency bands in input mel-spectrogram patch.
% EMBEDDING_SIZE = 128  # Size of embedding layer.

% # Hyperparameters used in feature and example generation.
% SAMPLE_RATE = 16000
% STFT_WINDOW_LENGTH_SECONDS = 0.025
% STFT_HOP_LENGTH_SECONDS = 0.010
% NUM_MEL_BINS = NUM_BANDS
% MEL_MIN_HZ = 125
% MEL_MAX_HZ = 7500
% LOG_OFFSET = 0.01  # Offset used for stabilized log of input mel-spectrogram.
% EXAMPLE_WINDOW_SECONDS = 0.96  # Each example contains 96 10ms frames
% EXAMPLE_HOP_SECONDS = 0.96 
From the raw audio recordings sampled at 16kHz sample rate, we compute log mel spectrograms with 64 frequency bands, a window length of 400 samples and hop size of 160 samples.
These spectrograms are then passed through a VGGish network \cite{VGGish} previously trained on the Audioset dataset, that generates one 128-dimensional embedding vector for each second of the log mel spectrograms. We use an openly-available pytorch implementation\footnote{https://github.com/harritaylor/torchvggish}. Additionally the generated embeddings are averaged over time in order to obtain a single embedding vector for each recording.
Before being fed into the network these embeddings are standardised based on the mean and standard deviation of the training set. 

The trainable network where the loss function is to be tested consists on a single linear layer that receives the 128 dimensional embedding and transforms it into a 3 dimensional embedding vector of norm 1. 
%  \vspace{-0.2cm}
\subsection{Evaluation}
\label{subsec:eval}

For evaluation purposes we want to focus on the quality of the embeddings learned, and how well these can express the hierarchical structure of the problems.
For that purpose we compute the silhouette score\cite{rousseeuw1987silhouettes} based on the ground truth labels. This score is computed by averaging the silhouette coefficient across all the samples in the set:%\footnote{from python package: sklearn.metrics.silhouette_score}}
\begin{equation}
    Sil =  \frac{1}{N}\sum_{n}^{N}\frac{b-a}{\max(a, b)}, 
    \label{eq:sil}
\end{equation}
Where $a$ is the intra-cluster distance and $b$ is the mean nearest-cluster distance.
This metric expresses how well the samples are positioned accordingly to the ground truth clusters, it has values ranging between $1$ and $-1$, from the best case where all samples are positioned within the correct cluster to the worst, where samples are 
positioned as if belonging to the wrong cluster. A value close to zero means that the clusters are overlapping and are difficult to separate.
For each set we compute the score using the labels at the fine level, and at the coarse level of the hierarchy, thus obtaining a measure of the quality of the learnt embeddings across the hierarchy. 

For the 3 datasets described before, we partition the pre-selected data into training set (70\%), validation set (20\%) and test set(10\%) at random.
Furthermore, for the majority of the experiments we use predefined batches that contain examples of pairs from all the ranks. This choice is based on the idea that balanced batches in terms of ranks would work better for the rank based loss, and it also makes comparisons with the quadruplet loss more complete.  %the quadruplets are constructed from the training and validation sets in separate, and because these are heavly constrained to follow the hierarchy, we seldom can use the full trianing and validation.
For all the experiments we employ an early stopping procedure (patience of 20) based on the average of silhouette scores in the validation set. \textit{i.e} the training stops once the max value for the averaged silhouette score is reached.
Furthermore, with the purpose of testing the capability of the models to generalise to unseen classes at the fine level, additional test sets were created from different fine classes than the development sets,
% \vspace{-0.3cm}
\subsection{Experiments}
% In this section we describe the experiments developed: % with the goal of testing the proposed loss function in different settings and understanding when this loss is more relevant.

\linesubsec{[InitEmb] Evaluation of initial pretrained embeddings} This serves the purpose of defining a baseline for comparison with all other experiments. The main goal is to understand if an improvement over the "non-hierarchical" initial embeddings is achieved or not. The embeddings dimensions were reduced to 3 in order to provide a fairer baseline for comparison. % the purpose is to understand if the "quality" of the initial embeddings impacts the performance of the Rank based loss.

\linesubsec{[QuadL] Quadruplet Loss} The  loss of  \cite{jati_hierarchy-aware_2019} is especially relevant to compare with the rank based loss. As  mentioned in sec.\ref{sec:intro}, this loss integrates hierarchical information through the  selection of the examples that generate the quadruplets. Batch size is 3 (quadruplets).

\linesubsec{[RbL] Rank based loss} Training the network with the RbL on the 3 datasets. Batch size is 12 and the examples are selected to create a balanced batch across ranks.

\linesubsec{[RbL\_unc] RbL with unconstrained batches} On all the previous experiments the batches are balanced regarding the hierarchical relationships between ground truth labels. Here that constraint is lifted, allowing, for example, batches to be missing pairs of one rank or have a disproportional number of pairs in another.

\section{RESULTS and DISCUSSION}
\label{sec:res}
Results are reported in Table \ref{tab:res}, giving the fine and coarse level silhouette scores obtained for the test sets. we also report the average between the silhouettes at both levels and to highlight how these compare with the baseline initial embeddings, the difference from the baseline is shown in brackets.

Generally, results show that both rank-based loss and quadruplet loss can learn to represent hierarchical embeddings, but with notable variation across datasets.
It is worth noting the difficulty in learning good embeddings to represent the fine level classes.
That aside RbL performs well including on novel leaf classes.

%Quadruplet loss results on normal test set:
Comparing the results for both test sets on the Nsynth and 3 bird datasets, they seem to show a general trend where when the initial representation is worse, RbL scores above the Quadruplet loss.
% A very low silhouette score for the initial embeddings at the fine level would indicate that the different classes within each coarse level categories are very similar and difficult to distinguish.
% We hypothesise that the positive result of the quad loss in this scenario is because of the margins that define a lower bound target distance between embeddings that is always different from zero.
% The RbL, not having these margins seems to be more influenced by the initial representation of the data, which given a particularly difficult case as such, will have more difficulty learning "good" embeddings. 
We hypothesise that this is related to the fact that Quadruplet loss defines a lower bound target distance between embeddings that is always different from zero and thus has less "liberty" when moving data around in the embedding space. 
% Thus scoring worse when it needs to modify the space a lot.
Another aspect of this is: by defining margins that are not based on the data, quadruplet loss will learn to position embeddings of different classes that are always distanced by the same amount. e.g, the distance between a flute, a  guitar and a keyboard are always the same. 
RbL however, since it gets the target distances from the data, could learn that a guitar and a keyboard are more acoustically close than a flute, and the learnt embeddings express this.   

% experiment for the future, does providing data that spans over a more complete hierarchy ( show more classes ) would work better? 
% the idea is that by doing so ...whattt?

% Or would this be achieved by simply giving it more examples per class?

Observing the results for experiment \textbf{[RbL\_unc]}, as expected the capability of learning hierarchical embeddings drops when we allow the batches to have any composition of examples regarding their hierarchical relationships. We argue however that this is an indication that RbL still allows some flexibility regarding the batch composition and that this is an advantage over the quadruplet loss that requires very strict quadruplet composition. The results for TUTasc2016 are an outlier in which the unconstrained batch performs better, an outcome which merits further study. 

% Finally, it is worth noting the results from table.\ref{tab:sils_hard_test_res}, these indicate the ability of RbL to learn "correct" embeddings for data of unseen fine level classes.

% \begin{table}[t]
%  \resizebox{\columnwidth}{!}{%
%  \centering
% \begin{tabular}{l|c|c|c|c|c|c|c|c|c}

%   & \multicolumn{3}{c|}{3 bird species} & \multicolumn{3}{c|}{Nsynth} & \multicolumn{3}{c}{TUTasc2016} \\ 
%   & $Sil^{Fine}$   & $Sil^{Coarse}$ & $avSil$  & $Sil^{Fine}$   &$Sil^{Coarse}$ & $avSil$  & $Sil^{Fine}$   & $Sil^{Coarse}$ &$avSil$ \\ \hline
% \textbf{InitEmb} & -0.17          & 0.49 &  0.16  & -0.14     & 0.11&  -0.02   & \textbf{0.7}    & 0.59&  \textbf{ 0.65}            \\ \hline
% \textbf{QuadL} &   -0.07 (+0.1)  &  \textbf{0.57 (+0.08)}& \textbf{0.25 (+0.09)}  & -0.09 (+0.05)   &    -0.02 (-0.13) & -0.06 (-0.04)        & 0.14(-0.56)   &     0.27 (-0.32) &   0.21 (-0.44)                             \\ \hline
% \textbf{RbL} &    \textbf{-0.05 (+0.12)}        &    0.48 (-0.01) & 0.22 (+0.06)   &        \textbf{-0.04 (+0.1)}     &\textbf{0.13 (+0.02)} & \textbf{0.05 (+0.07)}  &   0.15 (-0.55) &       \textbf{0.8 (+0.21)}&  0.48 (-0.17)      \\ \hline
% \textbf{RbL\_unc} &   -0.19 (-0.02)   &   0.32 (-0.17)  & 0.07 (-0.09) &  -0.33 (-0.19)  &     0.06 (-0.05)   & -0.14 (-0.12)  & 0.2 (-0.5)  & 0.74 (+0.15) & 0.47 (-0.18)       \\ \hline
% \end{tabular}%
% }
% \caption{Silhouette scores for the alternative test sets, where the leaf-classes are different from the classes used to train the models.}
% \label{tab:sils_hard_test_res}
% \end{table}

\vspace{-0.3cm}

\section{CONCLUSION}
\label{sec:conc}
This work presented a novel rank based loss function and we have shown its ability to learn embeddings that are representative of the hierarchy of the labels.
Our rank based loss was compared against another loss function that incorporates hierarchical information, with positive results.
Next steps involve a more in depth exploration regarding the effect of the dataset structure on the performance of RbL, the use of different distance metrics and evaluation of the approach from classification results.  Also it would be interesting to test the RbL with a larger number of hierarchical levels, and show its ability to deal with incomplete labelled data. 

% References should be produced using the bibtex program from suitable
% BiBTeX files (here: strings, refs, manuals). The IEEEbib.bst bibliography
% style file from IEEE produces unsorted bibliography list.
% -------------------------------------------------------------------------
\bibliographystyle{ieeebib}
%\bibliography{strings,refs}
\bibliography{main}

\end{document}